\documentclass[runningheads]{llncs}
\pdfoutput=1
\usepackage{graphicx}
\usepackage{comment}
\usepackage{amsmath,amssymb} 
\usepackage{color}
\usepackage{array}
\usepackage{diagbox}
\usepackage{multirow}
\usepackage{bm}

\usepackage[width=122mm,left=12mm,paperwidth=146mm,height=193mm,top=12mm,paperheight=217mm]{geometry}

\makeatletter
\def\thickhline#1{%
  \noalign{\ifnum0=`}\fi\hrule \@height #1 \futurelet
   \reserved@a\@xhline}
\makeatother

\begin{document}
\pagestyle{headings}
\mainmatter
\def\ECCVSubNumber{1348}  

\title{Distilling Object Detectors with Task Adaptive Regularization} 

\titlerunning{Distilling Object Detectors with Task Adaptive Regularization}
\authorrunning{Ruoyu Sun et al.}
\author{Ruoyu Sun$^{1\star}$, Fuhui Tang$^2$, Xiaopeng Zhang$^2$, Hongkai Xiong$^1$, and Qi Tian$^2$}
\institute{$^1$ Shanghai Jiaotong University\\ $^2$ Huawei Noah's Ark Lab}

\renewcommand{\thefootnote}{\fnsymbol{footnote}}
\footnotetext[1]{This work was done when the first author was an intern at Huawei Noah's Ark Lab.}
\maketitle

\begin{abstract}
Current state-of-the-art object detectors are at the expense of high computational costs and are hard to deploy to low-end devices. Knowledge distillation, which aims at training a smaller student network by transferring knowledge from a larger teacher model, is one of the promising solutions for model miniaturization. In this paper, we investigate each module of a typical detector in depth, and propose a general distillation framework that adaptively transfers knowledge from teacher to student according to the task specific priors. The intuition is that simply distilling all information from teacher to student is not advisable, instead we should only borrow priors from the teacher model where the student cannot perform well. Towards this goal, we propose a region proposal sharing mechanism to interflow region responses between the teacher and student models. Based on this, we adaptively transfer knowledge at three levels, \emph{i.e.}, feature backbone, classification head, and bounding box regression head, according to which model performs more reasonably. Furthermore, considering that it would introduce optimization dilemma when minimizing distillation loss and detection loss simultaneously, we propose a distillation decay strategy to help improve model generalization via gradually reducing the distillation penalty. Experiments on widely used detection benchmarks demonstrate the effectiveness of our method. In particular, using Faster R-CNN with FPN as an instantiation, we achieve an accuracy of $39.0\%$ with Resnet-50 on COCO dataset, which surpasses the baseline $36.3\%$ by 2.7\% points, and even better than the teacher model with 38.5$\%$ mAP.
\end{abstract}


\section{Introduction}\label{intro}
Object detection is a fundamental and challenging problem in computer vision. Various detection methods, varying from single-stage \cite{liu2016ssd,redmon2016you,sermanet2013overfeat} to two-stage \cite{girshick2014rich,girshick2015fast,ren2015faster}, have been proposed with the development of convolutional neural networks, and achieved significant improvement in accuracy. However, these approaches are usually equipped with cumbersome models and suffer expensive computation cost. Hence, designing neural networks with light computation cost as well as high performance, has attracted much attention in most real-world applications.
	
	
How to find the sweet spot between the accuracy and efficiency of a detection model has been explored for a long time. Knowledge Distillation (KD), introduced by Hinton \cite{hinton2015distilling}, has received much attention due to its simplicity and efficiency. The distilled knowledge is defined as soft label outputs from a large teacher network, which possibly contain the structural information among different classes. Following KD, many methods are proposed to either utilize the softmax outputs \cite{fukuda2017efficient,liu2018ktan} or mimic the feature layer of the teacher network \cite{romero2015fitnets,yim2017gift,zagoruyko2016paying}. However, these methods are mainly designed for multi-label classification, which cannot adapt to object detection directly. In principle, detection requires reliable localization in addition to classification, while the extreme imbalance of foreground and background instances makes it difficult to model intra-class similarities and inter-class differences. Moreover, object detection is a more complicated framework, especially for two-stage detectors \cite{ren2015faster,dai2016r,lin2017feature}, which combine several modules including region proposal network \cite{ren2015faster}, RoI pooling, and detection heads, \emph{etc.}. The interleaved relationships among them make it difficult to transfer knowledge directly.
	
	
To address the above issues, this paper proposes an adaptive distillation framework for a typical object detector, of which we deliberately design specific distillation strategy for each module according to their intrinsic properties. 
The highlight is that what we want to borrow from the teacher model is its generalization ability. Towards this goal, we propose a region proposal sharing mechanism to interflow regions between the teacher and student model, as to inspect its generalization ability over another model. The distillation operation is performed over regions where student model cannot perform well. In particular, our method adaptively mimics the responses of the teacher model in three aspects: 1) At the feature backbone level, we adaptively model the foreground regions for feature distillation. A two-dimensional Gaussian mask is introduced, emphasizing the target information while suppressing the irrelevant redundant background. As such, the more informative foreground is imitated more properly, which we find is the key for successful detection. 2) At the classification head level, benefiting from the region proposal sharing mechanism, the teacher model outputs soft labels for positive samples, which provides structural information among different classes as traditional knowledge distillation methods do. The student model is supervised by both soft labels and hard one-hot labels for better generalization. 3) At the bounding box regression level, the regressed bounding box locations from the teacher model are used as extra regressed targets for the student model. As such, the regression targets are progressively approaching to the ground truth, which we find is more robust for bounding box regression and achieve better generalization ability.

Equipped with the distillation loss, training the student model formulates a multi-task learning issue, where the distillation penalty enforces the model to imitate the teacher model, and the detection loss optimize the model as traditional detectors do. However, minimizing distillation loss and detection loss simultaneously would introduce optimization dilemma, where the optimal state suitable for distillation may not be applicable for detection. As a result, it is suboptimal when simply combining these two optimization targets. To solve this issue, we propose a distillation decay strategy to help improve model generalization via gradually reducing the distillation penalty. In this way, the distillation term can be treated as a guider to help the student model converge to a more better optimization point. Note that our method is a general framework, and can be adaptively incorporated into current widely used models (Faster R-CNN \cite{ren2015faster}, YOLO \cite{redmon2016you}, SSD \cite{liu2016ssd}) to improve the performance. In particular, using Faster R-CNN with FPN as an instantiation, we achieve an accuracy of $39.0\%$ with Resnet-50, which is even better than the teacher model with 38.5$\%$ mAP.
	
To sum up, this paper makes the following contributions:
	
$\bullet$ We propose a task adaptive distillation framework for object detection, which  adaptively transfers knowledge from teacher to student according to the task specific priors. This is achieved by imitating knowledge at three levels, \emph{i.e.}, feature backbone, classification head, and bounding box regression head, according to which model performs more reasonably.

$\bullet$ We propose a distillation decay strategy to gradually reduce the teacher's interference to the student, and help improve model generalization.

\section{Related Work}\label{related_works}
In this section, we review the literatures that are closely related to our work, including object detection and knowledge distillation.
\subsection{Object Detection}
The current mainstream CNN-based detection algorithms are mainly divided into single-stage and two-stage detectors. Single-stage object detectors such as YOLO \cite{redmon2016you} directly perform object classification and bounding box regression on the feature maps. SSD \cite{liu2016ssd} uses feature pyramid with different anchor sizes to cover the possible object scales. DSSD \cite{fu2017dssd} uses deconvolutional layers to build high-resolution feature maps with strong semantic information. RetinaNet \cite{lin2017focal} proposes focal loss to mitigate the unbalance between positive and negative examples. Two-stage detectors \cite{girshick2014rich,he2015spatial,girshick2015fast,ren2015faster,dai2016r} treat the detection as a `coarse-to-fine' progress via first generating candidate region proposals that contain the object of interests, and followed by a region refinement procedure for better localization. R-CNN \cite{girshick2014rich} and Fast R-CNN \cite{girshick2015fast} use proposals generated by object proposal algorithms, Faster R-CNN \cite{ren2015faster} introduces the Region Proposal Network (RPN) to produce region proposals. Cascade R-CNN \cite{cai2018cascade} uses multiple detection heads with increasing IoU thresholds to iteratively refine the detection results.

	
In order to accelerate the network training, some model compression techniques have been proposed. Weight quantization \cite{wu2016quantized,zhou2017incremental,han2015deep} refers to representing the weights with fewer bits, pruning approach \cite{han2015learning,park2016faster,tung2018deep} removes unimportant connections from a pre-trained network, and low-rank factorization \cite{sainath2013low,kim2015compression} exploits the redundancy in filters and feature map responses to simplify the model. However, these methods either change the network structure or have a lot of complexity, even hurt the performance significantly.
	
	\begin{figure*}[!t]
		\centering
		\includegraphics[width=1\textwidth, height=6cm]{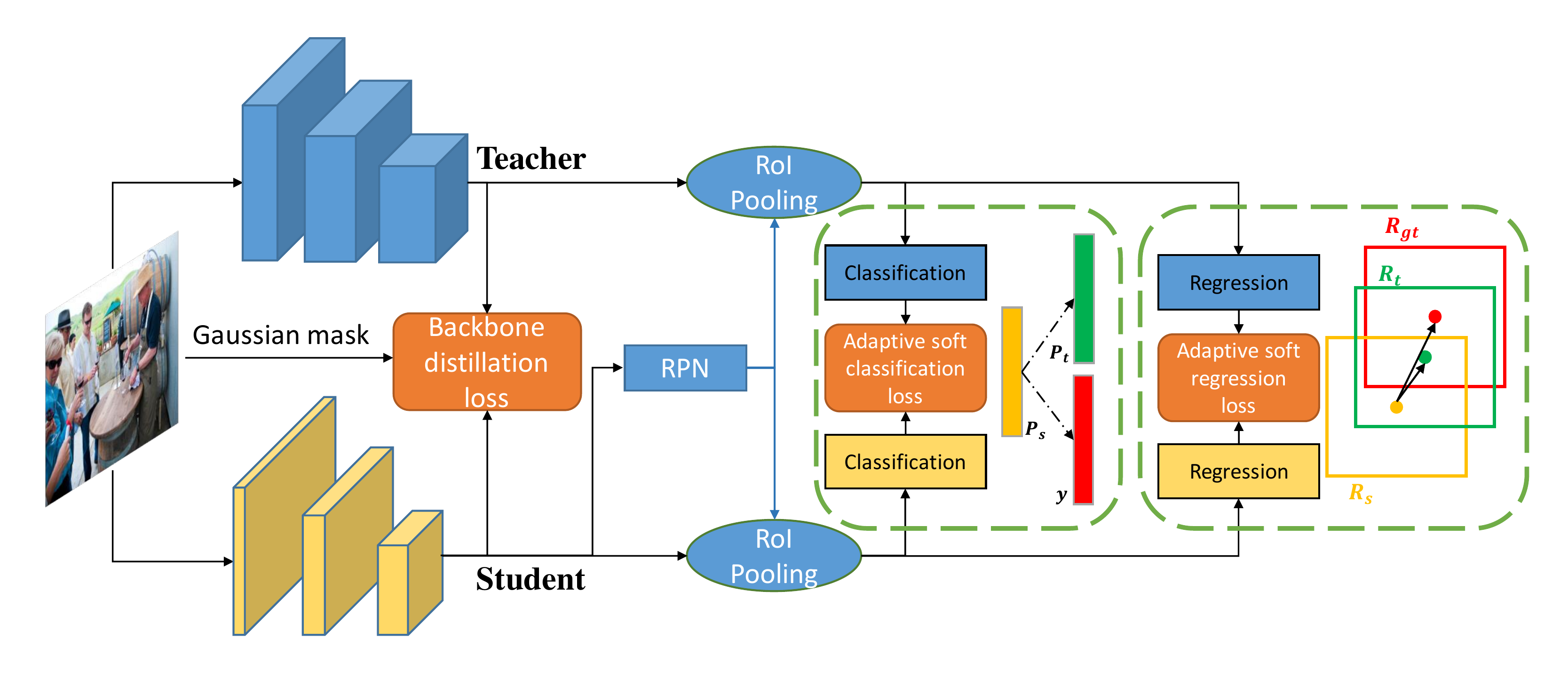}
\vspace{-1.0cm}
\setlength{\belowcaptionskip}{-0.4cm}
\caption{Overview of the proposed task adaptive distillation framework. The proposed distillation method consists of three modules: distillation of feature backbone, classification head, and bounding box regression head. Blue parts remain unchanged during distillation, while yellow parts are adaptively supervised by the teacher. Student's RPN is shared with teacher for imitation of classification and regression targets and their backbones' feature maps are imitated via an adaptive Gaussian mask around the foreground objects.}
		\label{figure2}
	\end{figure*}

\subsection{Knowledge Distillation}
Knowledge distillation is widely used to transfer knowledge from a high performance teacher model to a compact student model, aiming at improving the performance of the student model. It is first proposed by Hinton et al. \cite{hinton2015distilling} for image classification task, by utilizing the class probabilities as soft labels produced from the teacher model to guide the student's training. Hint learning \cite{romero2015fitnets} distills a deeper and thinner student model by imitating both the soft outputs and intermediate feature representations of the teacher model. Similar works are presented in \cite{zagoruyko2016paying,liu2018ktan,fukuda2017efficient} but are designed mainly for classifiers.
	
Recently, there are a few works that focus on detector distillation. Wang \emph{et al.} \cite{wang2019distilling} propose to distill the backbone with features on local regions near object, and have demonstrated the power for detector distillation. However, redundant selected background features limit the accuracy for distilling the most discriminative parts.  \cite{chen2017learning} tries to distill the student model from all components including feature maps, classification head and bounding box regression head but the imitation of entire feature maps also introduces too many backgrounds which would drown the positive samples. Li \emph{et al.} \cite{li2017mimicking} propose to transfer knowledge from both positive and negative region proposals on high-level features and corresponding task heads. Similar to \cite{chen2017learning}, it introduces too many background noise and are difficult for optimization. Meanwhile, \cite{li2017mimicking} makes student model imitate all of classification and regression results from teacher's task heads on every proposals simply by calculating L2 loss, which may cause misguidance when confronting proposals teacher model performs poorly on. 

In summary, current distillation frameworks either lack specific design for detectors or fail to effectively select the most critical part for distillation. Different from these works, our task adaptive distillation framework deliberately designs specific distillation strategy for each module according to their intrinsic properties and makes use of distillation decay strategy to further improve generalization.
	
\section{Method}\label{methodology}
\subsection{Network Overview}
In this section, we describe our proposed task adaptive distillation framework in detail. Without losing generality, we choose Faster R-CNN \cite{ren2015faster}, a typical two-stage object detector as an instantiation, and the whole framework is shown in Figure \ref{figure2}. The core idea of the proposed distillation method consists of three modules: distillation of feature backbone, classification head, and bounding box regression head, respectively. We adaptively transfer knowledge from teacher to student model with different imitation losses, based on the intrinsic property of each module.
	
In general, the distillation method is implemented via the proposed region proposal sharing mechanism, which interflow regions between the teacher and the student models to inspect how the candidate regions perform on another model. Based on the responses: 1) For backbone features distillation, we model the foreground regions with a two-dimensional Gaussian mask, emphasizing the target information while suppressing the redundant background noise. 2) For classification head, benefiting from the region proposal sharing mechanism, the teacher model outputs soft labels from classification head that uses the regions that student provides. In principle, these soft labels provide implied structural information beyond ground truth labels, and are able to achieve  better generalization when jointly trained. 3) For bounding box regression head, the regressed bounding box locations from the teacher model are used as extra regressed targets for the student model. As such, the regression targets are progressively approaching to the ground truth, which we find is more robust for bounding box regression and achieve better generalization ability. Each module would be elaborated as follows.
	
\subsection{Backbone Features}
For CNN-based detectors, the performance gain mainly credits to better backbone features. It is intuitive to mimic the backbone features of the teacher model for better performance. However, previous work \cite{wang2019distilling} has found that directly imitating features at all locations would not promote the performance much. The reason is that it is the foreground regions that counts for a successful detector. Since background regions especially easy negative regions usually occupy the majority of a feature map, merely perform distillation at all the locations would inevitably introduce large amount of noise from those uncared, easy negative regions.
	
	
	
\begin{figure*}[!t]
		\centering
		\includegraphics[width=1\textwidth]{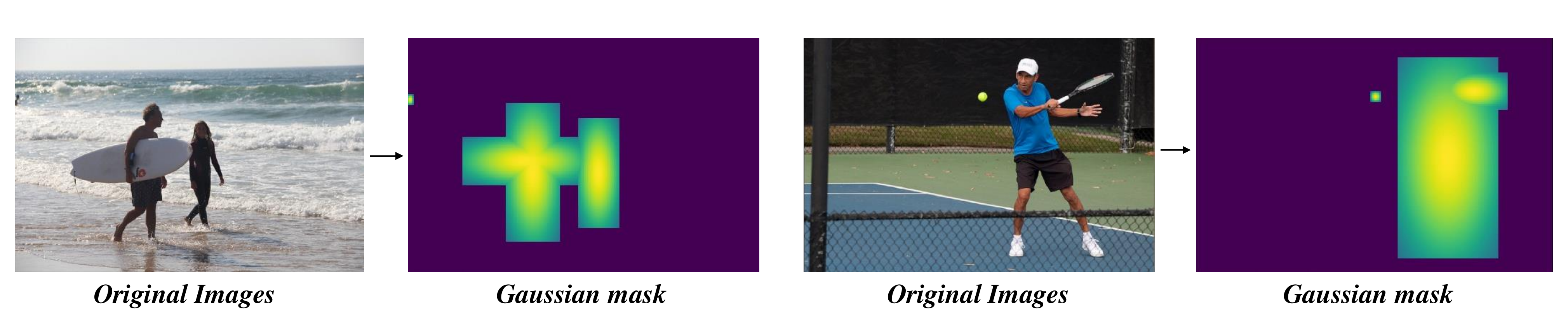}
\vspace{-0.8cm}
\setlength{\belowcaptionskip}{-0.45cm}
\caption{An illustration of the generated Gaussian masks over the COCO dataset.}
		\label{figure3}
	\end{figure*}

Based on the above observations, one intuitive strategy is to simply imitate the foreground ground truth regions. However, since the centric regions of an object are more likely to be sampled as positive samples by the following RoI layers, improving features around the centric regions would be better for model generalization. Hence, we introduce a Gaussian mask to highlight the centric foreground pixels and suppress those boundary regions around the objects. Specifically, given a bounding box $B$ of an object, with size of $w \times h$ and centered at $(x_0 , y_0 )$, the two-dimensional Gaussian mask is defined as:
	
\begin{equation}\label{mask_gaussian}
	M=\left\{
	\begin{array}{ll}
	e^{-\frac{{(x-x_0 )}^2}{\sigma_x^2{(w/2)}^2}-\frac{{(y-y_0)}^2}{\sigma_y^2{(h/2)}^2}}, & (x,y) \in B \\
	0, & \textrm{otherwise}
    \end{array}
	\right.
\end{equation}
	
where $\sigma_x^2$ and $\sigma_y^2$ are the decay factor along the two directions, and we set $\sigma_x^2=\sigma_y^2$ for simplicity. The mask is only effective within the ground truth bounding box and equals to zero everywhere else, thus we hope to only mimic the foreground object that we care about. In particular, when several Gaussian masks overlap over a single pixel $(x,y)$, the mask value is simply set as the maximum. Fig. \ref{figure3} illustrates some images, as well as the corresponding Gaussian masks.

With the delicately designed Gaussian mask, the backbone features are distilled via minimizing the following loss:
	
\begin{equation}\label{feature_loss}
	L_{bk}=\frac{1}{2N_{a}} \sum_{i=1}^W \sum_{j=1}^H \sum_{c=1}^C M^{ijc} (F_s^{ijc}-F_t^{ijc})^2,
\end{equation}
where $N_{a}= \sum_{i=1}^W \sum_{j=1}^H \sum_{c=1}^C M^{ijc}$ , $W$, $H$, $C$ are the width, height, and channels of the feature map, $F_s^{ijc}$ and $F_t^{ijc}$ denote the backbone features of student and teacher model, respectively.
	
\subsection{Classification Head}\label{Classification head distillation }
Knowledge distillation is widely used for classification task \cite{hinton2015distilling}, where the soft labels provided by teacher model contain structural information among different categories and can be regarded as some sort of regularization. However, directly transferring soft labels in a detection system is not applicable, mainly due to two aspects. First, the background regions occupy the majority of region proposals, which would introduce a large amount of noise if we consider their structural information among different classes; Second, the proposals output by two models are inevitably different, which makes them not comparable for knowledge distillation.
	
Our method effectively address the above issues for classification head distillation: 1) We do not rely on soft labels of the teacher model over those negative regions, instead we only focus on positive samples which are beneficial when modeling intra-class structural priors. 2) We propose to share the student's RPN proposals to the teacher model, and use the softened outputs as priors for distillation. Another advantage of the proposed region proposal sharing mechanism is that due to the large amount of possible region proposals (over millions of possible locations), the regions transferred from the student model would probably not seen by the teacher model, thus the classification output can be regarded as some sort of generalization priors for the student model to imitate.
	
Specifically, given $N$ region proposals from the output of the student's RPN layer, we compute the soft labels of the $N_p$ positive samples over the teacher model $\{y^i_t\}_{i=1}^{N_p} \in R^{C'}$, where $C'$ denotes the number of classes. Accompanying with all the proposals $N$ and their ground truth labels $\{y^j\}_{j=1}^{N} \in R^{C'}$, the loss for the classification head is reformulated as
	
\begin{equation}\label{loss_cls}
	L_{cls}= \sum_{j=1}^{N} L_{CE}(y^j_s, y^j)+\beta_1 \sum_{i=1}^{N_p} L_{BCE}(y^i_s,y^i_t),
\end{equation}
where $L_{CE}$ and $L_{BCE}$ denote cross-entropy and binary cross-entropy, respectively. $y_s$ is the prediction of the student model, and $\beta_1$ is a balancing factor that controls the two loss terms. By imitating the soft labels, the student model are able to learn hidden structural information of the objects.

\subsection{Bounding Box Regression Head}
As an important component in detection networks \cite{girshick2015fast,ren2015faster}, regression head is responsible for adjusting the location and size of a candidate region proposal, this is usually modeled as a smooth $L_1$ loss \cite{girshick2015fast}. In this formulation, the positive regions far away from the ground truth are usually assigned with more penalty. However, owing to the unbounded regression targets, directly raising the weight of localization loss of far away proposals will make the model more sensitive to outliers. These outliers, which can be regarded as hard samples, will produce excessively large gradients that are harmful for model convergence. As for bounding box regression distillation, we hope that the output of the teacher model could offer a reasonable mild target for the student model to regress, especially for those hard positive samples, and relieve the forced abrupt regression targets from current proposal to the ground truth.	
	
The teacher's regression output may provide wrong guidance for the student model, and even contradicts with the ground truth direction. Therefore, instead of using the teacher's regression output directly, we exploit an adaptive regression distillation strategy to selectively rely on teacher's outputs. Similar to classification head distillation, the student's RPN proposals are shared to the teacher model. Specifically, given $N_p$ positive region proposals from the student's RPN output, denote $\{{r}^i_{p}\}_{i=1}^{N_p}$, $\{{r}^i_t\}_{i=1}^{N_p}$, $\{{r}^i_s\}_{i=1}^{N_p}$, $\{{r}^i_{gt}\}_{i=1}^{N_p}$ as the bounding box locations before regression, teacher's regression output, student's regression output, and corresponding ground truth, respectively. We first calculate $IoU$ (Intersection-Over-Union) between ${r_p^i}$ and ${r_{gt}^i}$, and $IoU$ between ${r_t^i}$ and ${r_{gt}^i}$. If $IoU(r_t^i, r_{gt}^i)>IoU(r_p^i, r_{gt}^i)$, it indicates that the teacher's regression output is a reliable indicator to provide correct guidance for the student. Otherwise, we do not distill the student's regression head. The final regression loss is formulated as follows:
	
\begin{equation}\label{loss_reg}
	L_{reg}= \sum_{j=1}^{N} L(r^j_s,r^j_{gt})+\beta_2 \sum_{i=1}^{N_p} L_{dist}(r^i_s,r^i_t,r^i_{gt}),
\end{equation}
where $L$ is the smooth $L1$ loss defined as in \cite{ren2015faster}, $\beta_2$ is a balance factor, and $L_{dist}$ is the adaptive distillation loss:
	
\begin{equation}
	L_{dist}(r^i_s,r^i_t,r^i_{gt})=\left\{
	\begin{array}{ll}
	L(r^i_s,r^i_t) & \,\,\textrm{if}\,\,\,  IoU(r_t^i, r_{gt}^i)>IoU(r_p^i, r_{gt}^i) \\
	0. & \,\,\textrm{otherwise}
	\end{array}
	\right.
\end{equation}
	
Integrating the above three distillation terms produces our overall training targets of the student model,which can be formulated as:
	
\begin{equation}\label{total_loss}
	L = \lambda L_{bk} + L_{cls} + L_{reg} + L_{rpn},
\end{equation}
where $\lambda$ is the balance parameter for backbone distillation and $L_{rpn}$ is the RPN training loss in two-stage detector as described in \cite{ren2015faster}.

\subsection{Adaptive Distillation Decay}
The overall loss function in Eq. (\ref{total_loss}) simultaneously minimizes distillation loss and detection loss, which formulates a multi-task learning issue. However, the interleaved optimization among different targets would make training process unstable and hard to converge. The optimal state suitable for distillation may not be applicable for detection. To solve this issue, we propose a distillation decay strategy to help improve model generalization via gradually reducing the distillation penalty, hoping that the model focuses more on the detection task as the training process goes. This is achieved by introducing a time decay variable $\gamma(t)$, which decreases to 0 as the training proceeds. In our implementation, we simply set $\gamma(t) = 1-t/T$ at the $t$th training epoch, where $T$ is the total training epochs. The time decay variable is imposed to the balance parameter $\beta_1$, $\beta_2$, and $\lambda$ in Eq. (\ref{loss_cls}), Eq. (\ref{loss_reg}), Eq. (\ref{total_loss}) to control the intensity of distillation loss, \emph{i.e.,}

\begin{equation}\label{time_dacay}
	\tilde{\beta_1}=\gamma(t)\beta_1, \,\,\, \tilde{\beta_2}=\gamma(t)\beta_2, \,\,\, \tilde{\lambda}=\gamma(t)\lambda.
\end{equation}
	
\section{Experiments}
In this section, we evaluate our adaptive distillation framework for object detection, providing extensive designed evaluation and making comparison with state-of-the-arts.
\begin{table}[!t]
\setlength{\abovecaptionskip}{0.6cm}
\caption{Effects of various distillation modules on PASCAL VOC 2007.} \label{step_prob}
\fontsize{7.5pt}{11.25pt}\selectfont
\setlength\tabcolsep{6pt}
\center
\begin{tabular}{l|ccccccc}
Student R-50 &$\bm{\surd}$ & & & & & &  \\ \hline
Teacher R-101 & & & & & & & $\bm{\surd}$ \\ \thickhline{1pt}
Backbone? & &$\bm{\surd}$ & & & $\bm{\surd}$ & $\bm{\surd}$ & \\ \hline
Classification Head? & & & $\bm{\surd}$ & & $\bm{\surd}$ & $\bm{\surd}$ & \\ \hline
Regression Head? & & & &$\bm{\surd}$& $\bm{\surd}$& $\bm{\surd}$ & \\ \hline
Distillation Decay? & & & & & & $\bm{\surd}$ & \\ \hline
mAP (\%) & 70.0 & 72.4 & 73.2& 73.4 & 73.8 & 74.5 & 74.3\\ \hline
\end{tabular}
\vspace{-0.2cm}
\end{table}

\subsection{Experimental Setup}
\subsubsection{Datasets and Evaluation Metrics.} We evaluate our approach on two widely used detection benchmarks: 1) PASCAL VOC 2007 \cite{everingham2010pascal},  containing totally 9,963 images of 20 object classes, of which 5,011 images are included in \emph{trainval} and the rest 4,952 in \emph{test}; 2) Microsoft COCO \cite{lin2014microsoft}, a large scale dataset that contains over 135k images spanning 80 categories, of which around 120k images are used for \emph{train} and around 5k for \emph{val}. Following the default settings, for PASCAL VOC, we choose the \emph{trainval} split for training and the \emph{test} split for test, while for MS COCO, we choose the \emph{train} split for training and the \emph{val} split for test. For performance evaluation, the detection average precision(AP) is used, we report the COCO style (AP $[0.5:0.95]$) detection accuracy for MS COCO, and PASCAL style (AP $[0.5]$) accuracy for PASCAL VOC.
\subsubsection{Baseline Models.} We evaluate our method based on both two-stage and single-stage detection frameworks. For two-stage detectors, we choose widely used Faster R-CNN \cite{ren2015faster} detector and for single-stage detectors, we evaluate the performance using RetinaNet \cite{lin2017focal}. Since there are no RPN layers for RetinaNet and the anchors generated by teacher and student are exactly the same, we directly utilize the positive anchors for classification and regression head distillation. Besides, RetinaNet utilizes specially designed focal loss as the classification loss, so the distillation loss for classification head is also changed to focal loss. Other operations and parameters are the same with those in two-stage detectors. The Resnet \cite{he2016deep} series network are used as teacher and student, depending on their model size. For ease of narration, if Resnet-101 is used as teacher model, and Resnet-50 as student model, the distilled model is simply denoted as R-101-50.


\subsubsection{Implementation Details.} All experiments are performed on NVIDIA Tesla V100 8 GPUs with parallel acceleration. With Stochastic Gradient Descend (SGD) as optimization method, we set batch size to 16, allocating 2 images per GPU. The Gaussian parameters $\sigma_x^2$ and $\sigma_y^2$ in Eq. (\ref{mask_gaussian}) are set to 2. The balance factors $\beta_1$, $\beta_2$, and $\lambda$ are set as 10, 3, 0.6, respectively, via diagnosing the initial loss of each branch and ensuring that all losses are within the same scale. We find that these parameters are robust in our method, and do not affect the results too much as long as they are in the same scale. Unless specified, all experiments choose 1x learning rate for training 12 epochs. The resolutions for COCO, PASCAL VOC are set as (1333,800) and (1000,600), respectively, following the traditional implementation of each dataset.

\begin{figure*}[t]
	\centering
	\includegraphics[width=0.8\textwidth,height=110pt]{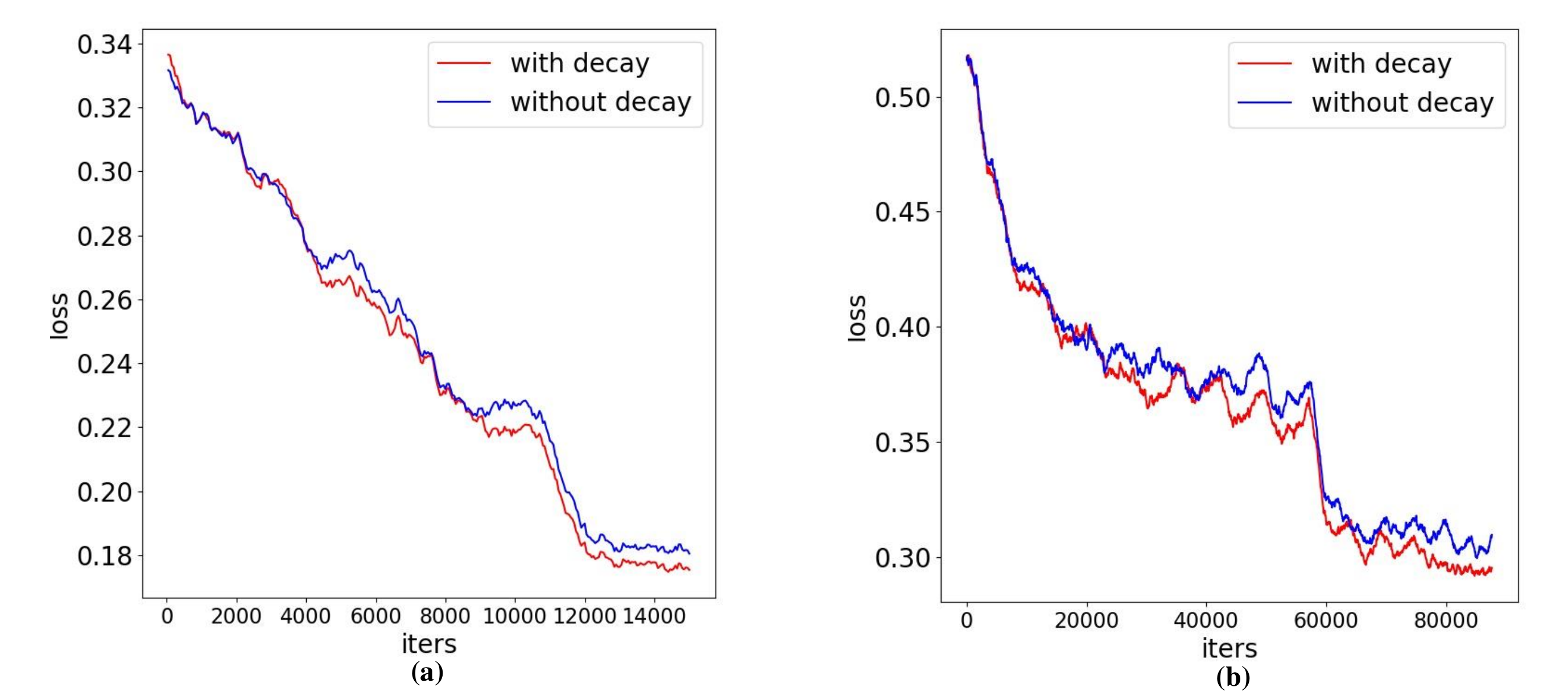}
	\setlength{\belowcaptionskip}{-0.3cm}
	\caption{The original training loss (including ROI loss and RPN loss) during the distillation procedure with and without distillation decay strategy. a) is calculated on VOC dataset, b) is calculated on COCO dataset. }
	\label{loss curve}
\end{figure*}

\subsection{Ablation Study}
We first conduct experiments to understand how each distillation module contributes to the final performance, as well as the robustness of our method to different parameters. Without loss of generality, all experiments in this section are based on PASCAL VOC 2007 with Resnet-101 as teacher and Resnet-50 as student, which produces accuracies of $74.3\%$ and $70.0\%$, respectively.

\subsubsection{Component Analysis.} We first conduct experiments with different configurations to reveal how each component affects the detection performance.  As is shown in Table \ref{step_prob}, different distillation components includes 1) backbone with Gaussian masks, 2) classification head, 3) regression head, 4) distillation decay. From the table we make the following observations:

\noindent $\bullet$ \emph{Backbone Distillation:} The backbone distillation brings about 2.4$\%$ mAP gain. The result demonstrates that our mask strategy enables the student model to learn highlighted foreground information.

\noindent $\bullet$ \emph{Classification and Regression Head Distillation:} The independent distillation strategies on classification head and regression head improve the student model by 3.2$\%$ and 3.4$\%$ points, respectively. It indicates that the classification head distillation can provide effective soft labels, while regression head gives the correct guidance.

\noindent $\bullet$ \emph{Combination:} The combination of the above three distillation targets achieves better results, which brings about another $0.4\%$ points gain compared with the best single distillation module. The combination strategy obtains marginal improvement compared with the individual components, partially due to the difficulty of joint optimization.

\noindent $\bullet$ \emph{Distillation Decay:} The distillation decay strategy can further improve the results from $73.8\%$ to $74.5\%$, which is even higher than the teacher model with an accuracy of $74.3\%$. This demonstrates the effectiveness of the proposed distillation decay strategy. As we decrease the guidance of the teacher, the student model is much easier for optimization. In this way, the teacher model can be treated as a guider that teaches the student model to reach to a better optimization point. Fig. \ref{loss curve} shows the training loss during the distillation procedure before and after introducing distillation decay strategy. As the training proceeds, the student model can emphasize more on the detection task itself and obtain lower training loss, which in turn improves the generalization ability of the model.

\begin{table}[t]
	\caption{Hyperparameter analysis of Gaussian mask's variances on PASCAL VOC 2007. `Rectangle' denotes using rectangle mask while `All features' denotes distilling the whole feature map.}
	\label{VOC_hyper}
	\begin{center}
	\fontsize{9pt}{13.5pt}\selectfont
	\setlength\tabcolsep{2.8pt}
     \setlength\tabcolsep{4pt}
		\begin{tabular}{l| *{4}{c|} c}
			\hline
			$\sigma_x^2$=$\sigma_y^2$ & 1 & 2 & 4 & Rectangle & All features\\
			\hline
			mAP & 73.1  & 73.2 & 73.1 & 72.7 & 72.1\\
			\hline
		\end{tabular}
	\end{center}
\vspace{-0.2cm}
\end{table}

\begin{table*}[t]
	\caption{Per category evaluation results on PASCAL VOC 2007.}
	\centering
	\fontsize{6pt}{11.25pt}\selectfont
	\setlength\tabcolsep{0.2pt}
	\begin{tabular}{l| *{20}{c|} c}
		\hline
		Model & mAP & aero & bike & bird & boat & bott. & bus & car & cat & chai. & cow & tabl. & dog & hors. & mbike & pern & plnt & sheep & sofa & train & tv\\
		\hline
		R-101 & 74.3 & 73.3 & 83.6 & 78.2 & 60.2 & 61.5 & 75.5 & 84.6 & 85.7 & 58.6 & 80.1 & 61.1 & 85.7 & 82.5 & 80.0 & 82.9 & 50.2 & 74.7 & 72.0 & 81.1 & 75.2\\
		R-50 & 70.0 & 67.1 & 79.1 & 73.9 & 56.3 & 54.4 & 74.9 & 81.5 & 82.7 & 51.7 & 76.4 & 53.2 & 82.3 & 81.4 & 75.7 & 80.5 & 45.0 & 71.9 & 64.0 & 77.7 & 70.9\\
		R-101-50 & 74.5 & 74.0 & 82.4 & 75.7 & 60.9 & 62.0 & 79.9 & 84.7 & 86.1 & 58.2 & 80.3 & 64.0 & 84.7 & 83.7 & 81.0 & 83.2 & 51.2 & 77.9 & 70.2 & 78.3 & 72.1\\
		\hline
	\end{tabular}
\vspace{0.1cm}
	\label{voc}
\end{table*}

\subsubsection{Hyperparameter of Gaussian Mask.}
We now investigate the influence of Gaussian mask for detection performance, which controls the imitation region of backbone features. For simplicity, we fixed $\sigma_x^2=\sigma_y^2$ for easy comparisons, and jointly change the two parameters by setting $\sigma_x^2 =\sigma_y^2=k$ to investigate the results. In principle, with larger $k$, the Gaussian mask becomes more scattered and more backgrounds are included, while with smaller $k$, the Gaussian mask becomes more concentrated on the center of the box, with less surroundings. As an extreme condition, when $\sigma_x^2=\sigma_y^2=+\infty$, the Gaussian mask degrades to a rectangle mask, which denotes the ground truth object bounding boxes. In order to verify the effectiveness of the Gaussian mask, we also take experiments on rectangle mask and the whole feature map (without any mask for regions selection). The results are  shown in Table \ref{VOC_hyper}, we find that the detection performance is relatively robust to the Gaussian mask and it is much better than simply using the ground truth rectangle or the whole feature maps for distillation. In particular, distilling all feature maps would deteriorate the performance, with $1.1\%$ points lower accuracy than using Gaussian masks. 

\begin{table*}[t]
	\caption{Distillation results of RetinaNet on MS COCO 2017, together with the model size and inference speed.}
	\label{retina-result}
	\begin{center}
	\fontsize{9pt}{13.5pt}\selectfont
	\setlength\tabcolsep{2.8pt}
		\begin{tabular}{l| *{6}{c|} c}
			\hline
			Network & Model info & mAP & AP50 & AP75 & APs & APm & APl\\
			\hline
			Retina-101 & 57.1M/10.9fps & 37.8 & 57.7 & 40.6 & 20.8 & 42.3 & 50.2\\
			\hline
			Retina-50 & \multirow{2}*{38.0M/12.1fps} & 35.6 & 55.2 & 38.1 & 20.0 & 39.3 & 47.4\\
			Retina-101-50 & ~ & 37.9 & 57.8 & 40.4 & 21.1 & 42.0 & 50.9\\
			\hline
		\end{tabular}
	\end{center}
\vspace{-0.1cm}
\end{table*}

\begin{table}[t]
	\caption{Evaluation results for different teacher and student models on MS COCO 2017.}
	\centering
	\fontsize{9pt}{13.5pt}\selectfont
	\setlength\tabcolsep{2.8pt}
	\begin{tabular}{l | *{6}{c|} c}
		\hline
		Network & Model info & mAP & AP50 & AP75 & APs & APm & APl\\
		
		\hline
		R-152 & 76.5M/10.8fps & 40.9 & 62.2 & 44.8 & 24.1 & 45.3 & 53.2\\
		\hline
		R-101 & \multirow{2}*{60.9M/11.9fps} & 38.5 & 60.2 & 41.7 & 22.5 & 42.9 & 49.2\\
		R-152-101 & ~ & 41.1 & 62.1 & 44.9 & 23.7 & 45.5 & 53.9\\
		\hline
		R-50 & \multirow{4}*{41.8M/13.6fps} & 36.3 & 58.0 & 39.2 & 21.6 & 39.9 & 46.4\\
		R-101-50 & ~ & 39.0 & 60.2 & 42.3 & 22.1 & 43.0 & 50.4\\
		R-152-50 & ~ & 39.9 & 61.0 & 43.6 & 22.8 & 44.0 & 52.5\\
		R-152-101-50 & ~ & 40.1 & 61.2 & 43.9 & 22.4 & 44.5 & 53.1\\

		\hline
	\end{tabular}
\vspace{0.1cm}

	\label{coco}
\end{table}

\subsection{Experimental Results}

\subsubsection{PASCAL VOC} The detection results by category on PASCAL VOC are shown in Table \ref{voc}. Our distillation model R-101-50 achieves a significant boost for each category, and brings $4.5\%$ overall gain (from $70.0\%$ to $74.5\%$) compared to the student model. It is surprising that the distillation result is even slightly better than the teacher model with accuracy of $74.3\%$, which demonstrates the superior performance of the proposed distillation framework.

\subsubsection{MS COCO} Table \ref{retina-result} and Table \ref{coco} show the overall distillation performance on COCO dataset for single-stage RetinaNet \cite{lin2017focal} and two-stage Faster R-CNN \cite{ren2015faster}, respectively. The Retina-101 improves Retina-50 by $2.3\%$ mAP and shows the superiority over the teacher Retina-101. Similarly, the distillation in Faster R-CNN also obtains outstanding performance. As is shown, the R-101-50 exceeds the teacher R-101 and the R-152-101 surpasses the teacher R-152. Specifically, the R-152-50 ($39.9\%$) even exceeds the R-101 ($38.5\%$) by a large margin. The reason for the gap between the R-152-50 and the teacher R-152 is that the feature difference between them is too large. Therefore, we explore a progressive distillation strategy to gradually reduce the gap. In detail, we first utilize R-101 as teacher model to distill R-50, and obtain R-101-50. Then we use the R-152 as teacher model to further distill R-101-50, the result model R-152-101-50 achieves a further mAP gain compared to the direct distillation strategy R-152-50. It is worth mentioning that although R-50 has much smaller layers than those of R-152, almost 1/3, our R-152-101-50 has a comparable result to the R-152.

\begin{figure*}[t]
	\centering
	\includegraphics[width=\linewidth]{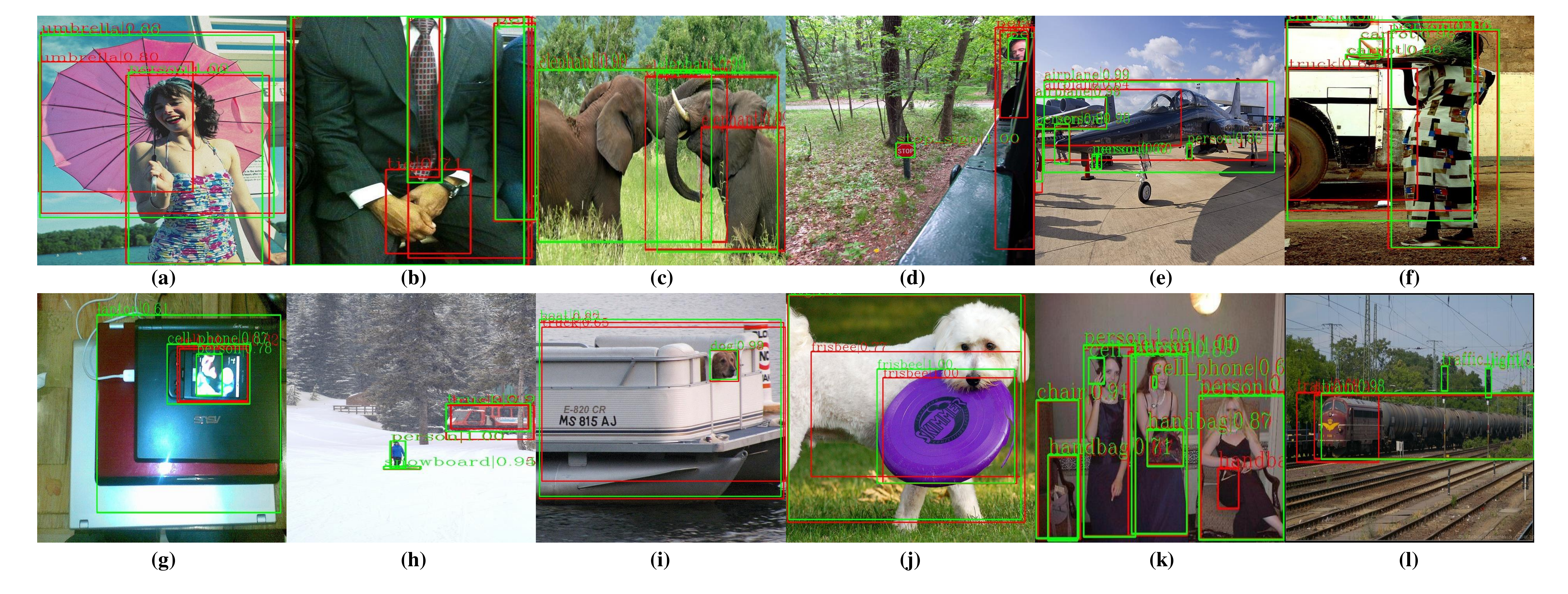}
	\setlength{\belowcaptionskip}{-0.3cm}
	\caption{Qualitative evaluation results on COCO 2017 dataset. The red and green bounding boxes denote detection results of student model before and after distillation, respectively. After distillation, some wrong detection results are suppressed and the positions of bounding boxes are more precise.}
	\label{visualization results}
\end{figure*}

For better understanding the detection performance, we visualize some detection results before and after distillation for qualitative evaluation. Figure \ref{visualization results} shows some detection results when using R-101 as teacher, R-50 as student, and Faster R-CNN as detector framework. We show some detection results before (red) and after (green) distillation. From these results we can observe that the performance of student is apparently propelled by the teacher model. Specifically, repeated detection, wrong proposals and grouped detection are successfully suppressed while bounding boxes around correct instances become more precise. 



\subsubsection{Compression and acceleration}
How to balance the speed and accuracy of a CNN model is always an open question. As is known, larger model usually offers higher accuracy while smaller model has higher speed. To better understand the compression and acceleration of knowledge distillation in object detection, we also provide the models' parameters amount and inference speed of each model in Table \ref{retina-result} and Table \ref{coco}.

The model's parameters amount and inference speed are shown in 'model info' column, where 'M' denotes 'million(s)' and 'fps' represents 'frames per second', respectively. According to the results, our distillation framework dramatically lightens the network size and increases the inference speed. For RetinaNet, Retina-101-50 compresses Retina-101 by about 68$\%$ with 1.2 fps gain. For Faster R-CNN, the distillation models (\emph{e.g.}, R-101-50, R-150-50) are much faster, while offering comparable or better performance against their teacher models.

\subsection{Comparison with Previous Distillation Methods}



To further explore the effectiveness of the proposed distillation framework in object detection, we present the comparative results between our method with two previous methods \cite{wang2019distilling,li2017mimicking}. For fair comparison, we uniformly adopt R-101 as the teacher model and R-50 as the student model. Since the devil is in the experimental details, the results of teacher and student models may differ in our implementation. We simply reimplement the results of \cite{li2017mimicking} which distills all proposals, while for method in \cite{wang2019distilling}, we simply choose the results in the original paper. From Table \ref{compare_SOTA}, we make the following observations: Mimic \cite{li2017mimicking} improves R-50 from $70\%$ to $72.7\%$ with $2.7\%$ mAP gain, while FGFI \cite{wang2019distilling} offers a little better result with $2.9\%$ mAP gain. Apparently, our approach outperforms these distillation methods on absolute performance and has a prominent mAP gain of $4.5\%$. The significant advantage of our method mainly comes from three aspects: 1) The Gaussian mask effectively suppresses the undesirable background information while retaining the target information; 2) Adaptive distillation for task heads provides suitable guide for the student model; 3) Distillation decay strategy helps for model optimization.


\begin{table}[t]
	\caption{Comparison with SOTA distillation methods for detectors on VOC 2007.}
	\label{compare_SOTA}
	\centering
	\fontsize{9pt}{13.5pt}\selectfont
	\setlength\tabcolsep{4.8pt}
	\begin{tabular}{c| *{3}{c|} c}
		\hline
		 & Network & Mimic\cite{li2017mimicking} & FGFI\cite{wang2019distilling} & Ours \\
		\hline
		Teacher & R-101 & 74.3 & 74.4 & 74.3\\
		Student & R-50 & 70.0 & 69.1 & 70.0\\
		Distilled & R-101-50 & 72.7 & 72.0 & \textbf{74.5}\\
		 &   & +2.7 & +2.9 & \textbf{+4.5}\\
		\hline
	\end{tabular}
\vspace{0.1cm}
\end{table}

\section{Conclusion}
In this paper, we proposed a task adaptive distillation framework for typical object detectors. The key contribution is that we deliberately design different imitating losses according to the property of each distilled target. Towards this goal, we propose a region proposal sharing mechanism to transfer regions between the teacher and student model, based on the responses, we are able to successfully select crucial part of teacher's feature maps, classification structural priors, and bounding box regression results as supervision for distillation. Besides, a distillation decay strategy is employed to help improve model generalization via gradually reducing the distillation penalty. Our distillation method is a general framework, and can be universally applicable for many modern detectors. Experiments conducted on widely used detection benchmarks demonstrate the effectiveness of the proposed method.

%

\clearpage
%
%
\bibliographystyle{splncs04}
\bibliography{manusript}
\end{document}